\documentclass[10pt,twocolumn,letterpaper]{article}

\usepackage{wacv}      

\usepackage{graphicx}
\usepackage{amsmath}
\usepackage{amssymb}
\usepackage{booktabs}

\usepackage[pagebackref,breaklinks,colorlinks]{hyperref}

\usepackage[capitalize]{cleveref}
\crefname{section}{Sec.}{Secs.}
\Crefname{section}{Section}{Sections}
\Crefname{table}{Table}{Tables}
\crefname{table}{Tab.}{Tabs.}

\def\wacvPaperID{*****} 
\def\confName{WACV}
\def\confYear{2025}

\begin{document}

\title{PoseGaussian: Pose-Driven Novel View Synthesis for Robust 3D Human Reconstruction}

\author{
Ju Shen$^{1}$ \quad
Chen Chen$^{2}$ \quad
Tam V. Nguyen$^{1}$ \quad
Vijayan K. Asari$^{1}$ \\
\\
$^{1}$University of Dayton \\
$^{2}$University of Central Florida \\
{\tt\small jshen1@udayton.edu, chen.chen@crcv.ucf.edu, tnguyen1@udayton.edu, vasari1@udayton.edu}
}
\maketitle

\begin{abstract}
  We propose PoseGaussian, a pose-guided Gaussian Splatting framework for high-fidelity human novel view synthesis. Human body pose serves a dual purpose in our design: as a structural prior, it is fused with a color encoder to refine depth estimation; as a temporal cue, it is processed by a dedicated pose encoder to enhance temporal consistency across frames. These components are integrated into a fully differentiable, end-to-end trainable pipeline. Unlike prior works that use pose only as a condition or for warping, PoseGaussian embeds pose signals into both geometric and temporal stages to improve robustness and generalization. It is specifically designed to address challenges inherent in dynamic human scenes, such as articulated motion and severe self-occlusion. Notably, our framework achieves real-time rendering at 100 FPS, maintaining the efficiency of standard Gaussian Splatting pipelines. We validate our approach on ZJU-MoCap, THuman2.0, and in-house datasets, demonstrating state-of-the-art performance in perceptual quality and structural accuracy (PSNR 30.86, SSIM 0.979, LPIPS 0.028).
\end{abstract}

\section{Introduction}
Novel View Synthesis (NVS) is a fundamental problem in computer vision and graphics, crucial for applications such as virtual reality and telepresence. It has evolved from traditional image-based rendering (IBR) techniques~\cite{Pons2007MultiView, Zhou2018, flynn2016deepstereo} to advanced neural representations like Neural Radiance Fields (NeRF)~\cite{mildenhall2021nerf} and their extensions~\cite{pumarola2021dnerf, tretschk2021nonrigid, zhang2021editable}, achieving impressive realism in view generation.
This advancement has sparked growing interest in free-viewpoint human rendering, an important NVS subfield. Nevertheless, synthesizing high-quality human views remains difficult due to dynamic motion, individual appearance variations, and real-time constraints. Recent works have adapted NeRF frameworks to capture human appearance and motion \cite{peng2021neural, su2021anerf}. Although effective in modeling person-specific details, the implicit volumetric rendering approach—which requires dense spatial sampling—creates a bottleneck for real-time applications.

\begin{figure*}[t]
  \centering
  \includegraphics[width=\textwidth]{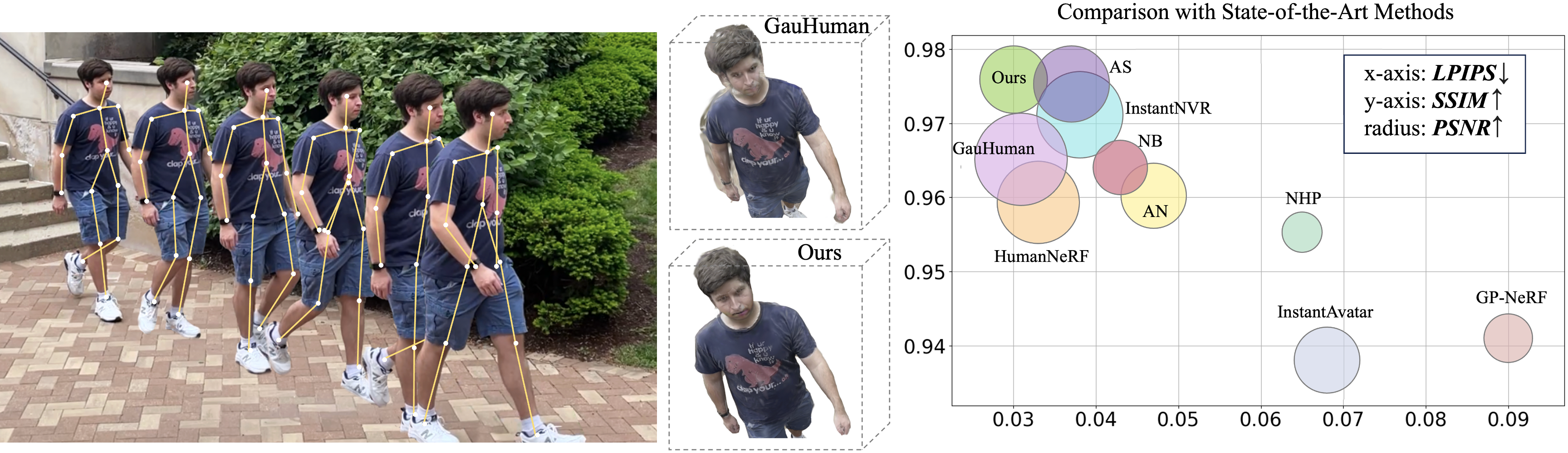}
  \caption{\textbf{PoseGaussian Visualization and Comparison.} Left: reconstructed pose-guided motion sequence reprojected into the original scene. Middle: visual comparison with GauHuman~\cite{hu2024gauhuman}. Right: performance chart compared to selected methods on the three metrics LPIPS, SSIM, PSNR. Compared methods: GauHuman (~\cite{hu2024gauhuman}), AS~\cite{zhou2024animatable}, InstantNVR~\cite{geng2023learning}, HumanNeRF~\cite{Zhao_2022_CVPR}, NB~\cite{peng2023implicit}, AN~\cite{peng2021animatable}, NHP~\cite{kwon2021neural}, InstantAvatar~\cite{jiang2023instantavatar}, GP-NeRF~\cite{chen2022gpnerf}.}
  \label{fig:teaser}
\end{figure*}

Recent advances in 3D Gaussian Splatting (3D-GS) have demonstrated impressive runtime performance while maintaining high visual fidelity~\cite{kerbl2023gaussian}. Unlike NeRF, 3D-GS adopts an explicit functional representation that enables direct projection of Gaussian primitives, making it well-suited for real-time applications. This paradigm has shown strong potential in domains such as free-viewpoint human rendering~\cite{zheng2024gpsgaussian, hu2024gauhuman, kocabas2024hugs}. Despite these successes, existing methods largely rely on low-level appearance cues and struggle to handle complex human motions. For instance, while GPS-Gaussian~\cite{zheng2024gpsgaussian} demonstrates real-time human view synthesis, it does not incorporate explicit human semantics—such as pose or body structure—into the prediction of Gaussian parameters. On the other hand, GauHuman~\cite{hu2024gauhuman} introduces structured human priors into the pipeline by leveraging fits from SMPL (Skinned Multi-Person Linear Model)~\cite{loper2015smpl} to guide Gaussian generation. However, its strong reliance on accurate SMPL alignment limits generalization in the presence of pose noise or when body shapes deviate from the training distribution. Moreover, it struggles with rapid motion, often producing motion blur and background leakage. 

To bridge the gap between low-level Gaussian representations and the effective integration of human priors, we introduce a pose-guided 3D-GS representation that incorporates body pose and temporal dynamics directly into the rendering pipeline. This structured prior provides strong semantic anchoring and enhances temporal coherence, particularly under sparse views or rapid motion.  Figure~\ref{fig:teaser} (left) illustrates the core concept of our method—reconstructing a pose-guided motion sequence and reprojecting it into the scene—showing how extracted pose priors guide coherent temporal updates. Figure~\ref{fig:teaser} (right) presents a quantitative comparison against recent state-of-the-art methods on the standard public benchmark ZJU-MoCap~\cite{peng2021neural}, where our approach consistently achieves superior performance across the three metrics—PSNR~\cite{huynh2008scope}, SSIM~\cite{wang2004image}, and LPIPS~\cite{zhang2018unreasonable}. To emphasize the perceptual sharpness and structural fidelity of our results, Figure~\ref{fig:teaser} (middle) visualizes reconstructed novel views alongside GauHuman~\cite{hu2024gauhuman}, which exhibits motion blur and unclear boundaries under dynamic motion. Our main contributions are as follows:

\begin{itemize}
    \item \textbf{Pose-aware initialization scheme} that leverages human body priors to guide the estimation of Gaussian primitives, enabling semantically informed and structurally coherent representations.
    \item \textbf{Temporal regularization strategy} that promotes inter-frame consistency while preserving fine-grained details and accommodating natural motion variations.
    \item \textbf{End-to-end differentiable framework} that jointly incorporates pose-informed feature encoding and motion-consistent regularization within a unified architecture for dynamic human reconstruction.
\end{itemize}

Our approach achieves state-of-the-art performance on several benchmark datasets, demonstrating superior reconstruction quality, enhanced temporal stability, and real-time rendering efficiency compared to existing methods. 

\section{Related Works}
Human view synthesis has evolved from early 3D reconstruction using structured light and laser scanning, to model-driven methods like SCAPE \cite{anguelov2005scape} and SMPL \cite{loper2015smpl}, which represent human bodies with articulated models. Though influential, these methods often lacked photorealism and required strong priors or manual tuning. More recently, learning-based techniques such as Neural Radiance Fields (NeRF) \cite{mildenhall2021nerf} have enabled photorealistic novel view synthesis directly from images. Building on this, 3D-GS \cite{kerbl2023gaussian} introduces an explicit, point-based representation using 3D Gaussians to improve rendering efficiency. We focus on this modern paradigm for human appearance reconstruction.

\begin{figure*}[t]
  \centering
  \includegraphics[width=0.8\linewidth]{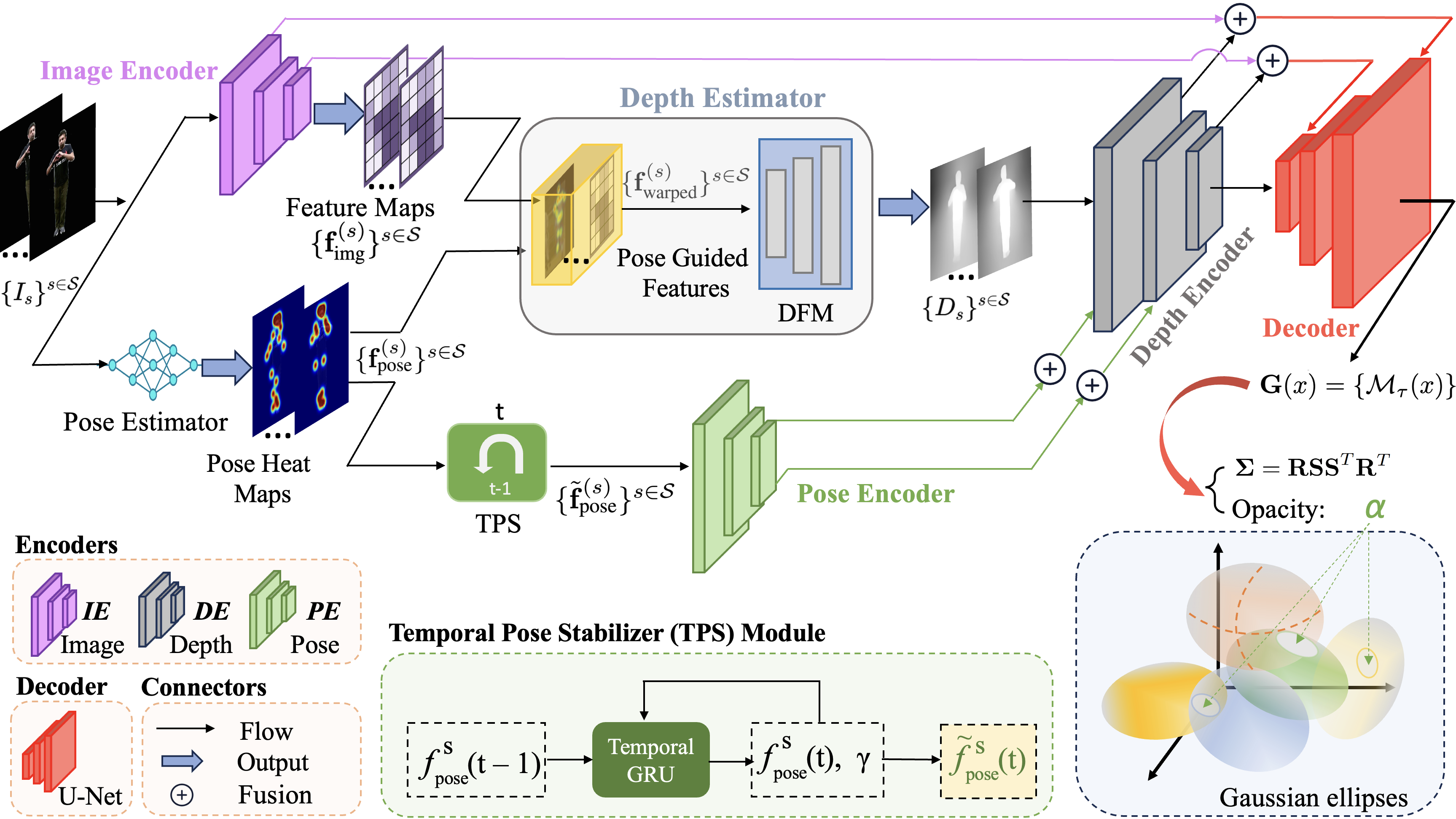}
\caption{The PoseGaussian pipeline. \textit{Top:} The overall workflow, illustrating the process from input color images to the predicted Gaussian parameter maps, specifically the rotation  $\mathcal{M}_r(x)$, scale $\mathcal{M}_s(x)$, and opacity  $\mathcal{M}_\alpha(x)$. \textit{Bottom:} A detailed view of the Temporal Pose Stabilizer (TPS) module, along with visual annotations clarifying the roles of various modules and connections.
}
  \label{fig:pipeline}
\end{figure*}

\textbf{Radiance Fields and Human Representation:} NeRF has been extended to human subjects with minimal motion \cite{park2021nerfies, park2021hypernerf, li2021nsff}, enabling view synthesis without explicit motion modeling. Several methods incorporate parametric human models like SMPL \cite{loper2015smpl} or 3D skeletons \cite{zhang2021editable, peng2021neural, peng2021animatable} to support pose-conditioned rendering. Neural Body \cite{peng2021neural} and Neural Actor \cite{liu2021neuralactor} apply sparse convolutions and mesh-based warping, but often require per-subject optimization and generalize poorly to unseen poses. Others, such as HumanNeRF \cite{weng2022humannerf} and Vid2Actor \cite{weng2020vid2actor}, focus on single-video rendering, yet struggle to capture fine details without dense supervision. Dynamic scenarios further expose limitations in pose estimation and warping, often resulting in artifacts like flickering and jittering \cite{liu2021neuralactor, peng2021animatable}. Overall, implicit volumetric NeRF models remain computationally heavy and memory-intensive, with slow training and inference. While acceleration techniques exist \cite{fridovich2022plenoxels, hedman2021baking}, the implicit nature still impedes scalability and real-time use.

\textbf{Gaussian Splatting for Human Novel View Synthesis:}
3D-GS \cite{kerbl2023gaussian} has quickly become a leading direction for human-centric representations, sparking a wave of influential work \cite{kocabas2024hugs, moreau2024human, Pang_2024_CVPR, hu2024gauhuman, li2024gaussianbody, qian20233dgsavatar, shao2024splattingavatar}. One growing line of research explores sparse-camera setups to improve efficiency and training speed \cite{zheng2024gpsgaussian, HFGaussian24}, which are promising for real-time applications but may struggle with capturing fine-grained geometry or detailed motion. For example, GPS-Gaussian \cite{zheng2024gpsgaussian} omits explicit human modeling, while HFGaussian \cite{HFGaussian24} relies on depth maps predicted from RGB, potentially introducing temporal artifacts. Another prominent direction combines Gaussian Splatting with parametric body models like SMPL, typically using Linear Blend Skinning (LBS) \cite{kavan2007skinning}, to achieve more accurate and controllable dynamic human representations \cite{kocabas2024hugs, moreau2024human, Pang_2024_CVPR, shao2024splattingavatar, li2024gaussianbody, hu2024gauhuman}. While this improves realism, it often depends on fixed body templates and complex pipelines, limiting generalization across diverse identities and motion types \cite{kocabas2024hugs, shao2024splattingavatar}. Our method bridges both directions with a lightweight, model-free design that introduces human structure via sparse 3D pose cues, retaining efficiency while improving stability and adaptability across varied motions and appearances.

\section{The PoseGaussian Method}
\label{label:method}
Fig.~\ref{fig:pipeline} illustrates the overall architecture, which comprises three primary encoders: an image encoder (\textit{IE}) for extracting dense visual features, a pose encoder (\textit{PE}) for interpreting skeletal keypoint distributions, and a depth encoder (\textit{DE}) for estimating scene geometry. The pipeline takes input images processed by the IE and fuses pose heatmaps from a pre-trained estimator with image features, providing the DE with human-centric cues to improve scene geometry. In parallel, the keypoint heatmaps are further processed through a Temporal Pose Stabilizer (TPS) module and subsequently encoded by the PE. The resulting pose features are decoded with the help of joint skip connections to reinforce temporal consistency and preserve semantic details. The final output consists of pixel-aligned 2D Gaussian parameter maps $\mathbf{G}(x) = { \mathcal{M}_\tau(x) }$, where $\tau \in {\text{p}, \text{c}, \text{r}, \text{s}, \alpha}$ denotes the projected position, color, rotation, scale, and opacity at each pixel location $x$. While 3D Gaussians provide stronger view consistency, our choice of 2D primitives avoids explicit 3D reconstruction and allows for more flexible modeling of image-space variations, which aligns better with our pose-conditioned, mesh-free formulation.

\subsection{Pose-Guided Feature Fusion for Depth Inference}
\label{lable:depthgeneration}

To improve geometric accuracy under motion, we leverage pose priors that inform the visual representation by introducing semantically aligned structural guidance. Simple pose heatmaps—such as those from BlazePose~\cite{bazarevsky2020}—are lightweight, resolution-aligned, and readily compatible with the \textit{PE}, making them well-suited for this purpose. Our method is compatible with various pose estimation backbones: when pose data is available in different formats (e.g., from prior works on robust pose estimation~\cite{liu2020attention, zheng20213d}), joints can be projected and encoded as 3D Gaussian heatmaps via joint-to-heatmap encoding. While our current implementation uses RAFT-Stereo for improved depth in binocular setups, PoseGaussian can also be adapted to monocular sequences by substituting the depth module with monocular estimators or omitting depth cues entirely, enabling pose-conditioned generation from single-view sources~\cite{qu2022heatmap}. Future work will examine how the accuracy and nature of the pose estimation backbone affect final reconstruction quality.


From each source view \( s \in \mathcal{S} \), two types of features are obtained: image features are extracted using a convolutional encoder, while pose features are generated directly by a pose estimation network \cite{bazarevsky2020}.
\begin{equation}
\mathbf{f}^{(s)} =
\left\{
\begin{array}{l}
\mathbf{f}^{(s)}_{\text{img}} = \mathcal{E}_{\text{img}}(I_s) \in \mathbb{R}^{H/2^k \times W/2^k \times D_i}, \\
\mathbf{f}^{(s)}_{\text{pose}} = P_s \in \mathbb{R}^{H/2^k \times W/2^k \times D_p}
\end{array}
\right.
\label{eq:poseandfeature}
\end{equation}
where \( P_s \) and \( I_s \) are the 2D pose and RGB image from the \( s \)-th source view. The spatial resolution \( H/2^k \times W/2^k \) reflects the output size after \( k \) stages of downsampling in the encoder, with \( k = 3 \) or \( 4 \) in our implementation. Here, \( D_i \) and \( D_p \) denote the channel dimensions for the image and pose features, respectively. The two streams are fused along the channel dimension to form the unified feature map \( \mathbf{f}^{(s)} \), where simple concatenation or alternative fusion strategies (e.g., \cite{hou2019deep, song2009novel}), as discussed in the ablation study. The fused feature set \( \{\mathbf{f}^{(s)}\}^{s \in \mathcal{S}} \) is first warped to the target view using the corresponding extrinsics and estimated depth from the DE module, producing the warped feature representation \( \{\mathbf{f}_{\text{warped}}^{(s)}\}^{s \in \mathcal{S}} \). Following the design of RAFT-Stereo~\cite{lipson2021raftstereo}, we construct a correlation-based feature volume by aggregating view-wise similarities between the warped source features \( \mathbf{f}_{\text{warped}}^{(s)} \) and the target view \( \mathbf{f}_{\text{warped}}^{(t)} \):
\begin{equation}
\mathbf{C}_{ijk} = \sum_{s \in \mathcal{S}} \sum_h \mathbf{f}^{(t)}_{\text{warped}}(i,j,h) \cdot \mathbf{f}^{(s)}_{\text{warped}}(i,k,h)
\end{equation}
Here, \( (i, j, h) \) and \( (i, k, h) \) represent spatial locations and feature channels, with \( h \) indexing the concatenated color and depth features, and \( j \) and \( k \) indexing different spatial positions. The resulting correlation features form a 3D volume \( \mathbf{C} \), which is then fed into a lightweight GRU-based update module, referred to as the \textit{DFM} (Depth Refinement Module) in Fig.~\ref{fig:pipeline}, that iteratively refines the depth prediction~\cite{lipson2021raftstereo}.
\begin{equation}
D_s^{(t)} = \Phi_{\text{depth}}(\mathbf{C}, \{K_s\}_{s \in \mathcal{S}}; D_s^{(t-1)}), \quad t = 1, \dots, T
\end{equation}
where $D_s^{(t)}$ denotes the estimated depth map for source view $s$ at iteration $t$, $\Phi_{\text{depth}}$ is a lightweight GRU-based update module, and $\{K_s\}_{s \in \mathcal{S}}$ are the camera intrinsics of the source views. At each iteration, the module refines the depth estimate using the previous prediction $D_s^{(t-1)}$ and local correlation slices extracted from the cost volume $\mathbf{C}$.

\subsection{Temporal Pose Stabilization for Feature Guidance}
\label{lable:temporalStabilization}
Beyond depth generation, pose features play a crucial role in enhancing the decoding stage for Gaussian parameter prediction. As illustrated in Fig.~\ref{fig:pipeline}, the heatmap output $\{\mathbf{f}^{(s)}_{\text{pose}}\}^{s\in\mathcal{S}}$ branches into a secondary pathway directed to the pose decoder, whose feature maps are then fused with the corresponding outputs from the depth and image encoders. Together, these fused features form the skip connections (indicated by red arrows) that feed into the decoder, which adopts a U-Net-style architecture to predict per-pixel Gaussian attributes, including rotation map $\mathcal{M}_r$, scale map $\mathcal{M}_s$, and opacity map $\mathcal{M}_\alpha$. Pose heatmaps serve as a dynamic auxiliary stream, injecting semantic human structural cues into skip connections to reinforce encoder-decoder alignment and preserve structural consistency during occlusion and fine-grained motion. To ensure reliable pose-based guidance in dynamic scenes, we introduce a \emph{Temporal Pose Stabilizer (TPS)} module that processes heatmaps across adjacent frames. Without such filtering, temporal inconsistencies in pose can lead to flickering artifacts, a common issue in prior methods. Inspired by temporal filtering techniques in video-based pose estimation~\cite{NIPS2019_gberta, Feng_2023_CVPR}, TPS employs a lightweight recurrent mechanism to smooth pose signals and suppress jitter caused by motion or occlusion. This is particularly important for dynamic, non-rigid subjects like humans, where even minor inconsistencies in pose estimates can degrade reconstruction quality (see our ablation study for analysis). Specifically, TPS takes the pose heatmaps from two consecutive frames, denoted as $\mathbf{f}^{(s)}_{\text{pose}}(t)$ and $\mathbf{f}^{(s)}_{\text{pose}}(t-1)$, and applies a temporal filter to obtain a smoothed pose signal $\tilde{\mathbf{f}}^{(s)}_{\text{pose}}(t)$:
\begin{equation}
\tilde{\mathbf{f}}^{(s)}_{\text{pose}}(t) = \omega \cdot \mathbf{f}^{(s)}_{\text{pose}}(t) + (1 - \omega) \cdot \mathbf{f}^{(s)}_{\text{pose}}(t-1)
\end{equation}
where $\omega \in [0, 1]$ is a blending factor that controls the contribution of the current frame and the previous frame’s pose signal. We use the past frame \( (t-1) \) instead of the future frame \( (t+1) \) to ensure causality, which is essential for real-time inference and streaming scenarios. This recurrent mechanism ensures that the pose representation is temporally stable and smooth across frames. Importantly, TPS operates as a pre-processing step before pose encoding, preserving the overall structure of the feature fusion and depth estimation modules.

\subsection{Pose-Conditioned Loss Objective}

Our training objective integrates pose-aware supervision into the standard differentiable rendering framework, enhancing both geometric fidelity and human-centric feature guidance. The overall loss function is defined as:
\begin{equation}
\mathcal{L}_{\text{total}} = \mathcal{L}_{\text{render}} + \mathcal{L}_{\text{depth}} + \mathcal{L}_{\text{pose-fusion}}
\end{equation}
While prior works~\cite{Pang_2024_CVPR, li2024gaussianbody, zheng2024gpsgaussian} typically optimize only for image reconstruction and depth consistency, our formulation explicitly introduces a pose-conditioned feature alignment loss, enabling stronger structural supervision during training. Specifically, the photometric rendering loss $\mathcal{L}_{\text{render}}$ combines pixel-level fidelity and structural similarity between the rendered view $\hat{I}$ and ground-truth $I$:
\begin{equation}
\mathcal{L}_{\text{render}} = \beta \, \mathcal{L}_{\text{MAE}}(\hat{I}, I) + \gamma \, \mathcal{L}_{\text{SSIM}}(\hat{I}, I)
\label{eq:loss1}
\end{equation}
where $\beta$ and $\gamma$ balance the reconstruction terms. The depth supervision term $\mathcal{L}_{\text{depth}}$ follows an exponentially weighted scheme to encourage consistent depth predictions across multiple decoding stages:
\begin{equation}
\mathcal{L}_{\text{depth}} = \sum_{t=1}^{T} \mu^{T-t} \, \left\| d_t - d_{\text{gt}} \right\|_1
\end{equation}
where $d_t$ denotes the predicted depth at stage $t$, $d_{\text{gt}}$ is the ground-truth depth, and $\mu$ controls the decay rate. Finally, the pose-fusion loss \( \mathcal{L}_{\text{pose-fusion}} \) supervises intermediate feature decoding by aligning the fused features \( f_{\text{joint}} \)—which result from integrating both visual and pose cues—with the pose-encoded features \( f_{\text{pose}} \):
\begin{equation}
\mathcal{L}_{\text{pose-fusion}} = \lambda \, \left\| f_{\text{joint}} - f_{\text{pose}} \right\|_1
\label{eq:loss2}
\end{equation}
This term ensures that essential human-centric structural priors, captured by the pose encoder, are preserved throughout decoding. By enforcing this alignment, the model learns to retain and propagate pose-aware semantics, enhancing spatial reasoning in the 3D reconstruction.

\section{Experiment Results}

\begin{table*}[t]
  \centering
  \resizebox{\textwidth}{!}{%
  \begin{tabular}{lcccc|lccc}
    \toprule
    \multicolumn{5}{c|}{\textbf{ZJU-MoCap}} & \multicolumn{4}{c}{\textbf{Twindom}} \\
    Method & PSNR↑ & SSIM↑ & LPIPS↓ & Train/FPS & Method & PSNR↑ & SSIM↑ & LPIPS↓ \\
    \midrule
    PixelNeRF [CVPR'21]~\cite{yu2021pixelnerf} & 24.71 & 0.892 & 0.120 & 1h / 1.20 & KeypointNeRF [ECCV'22] ~\cite{mihajlovic2022keypointnerf} & 19.68 & 0.890 & - \\
    NHP [NIPS'21]~\cite{kwon2021neural} & 28.25 & 0.955 & 0.065 & 1h / 0.15 & PixelHuman [arXiv'23]~\cite{shim2023pixelhuman} & 24.20 & 0.948 & - \\
    NB [CVPR'21]~\cite{peng2021neural} & 29.03 & 0.964 & 0.043 & 10h / 1.48 & 3D-GS [ACMTOG'23]~\cite{kerbl2023gaussian} & 22.77 & 0.785 & 0.153 \\
    AN [CVPR'21]~\cite{peng2021animatable}  & 29.77 & 0.965 & 0.470 & 10h / 1.11 & FloRen [SIGGRAPH'22]~\cite{shao2022floren}  & 22.96 & 0.838 & 0.165 \\
    AS [TPAMI'24]~\cite{zhou2024animatable}  & 30.38 & 0.975 & 0.037 & 10h / 0.40 & IBRNet [CVPR'21]~\cite{Wang_2021_CVPR} & 22.92 & 0.803 & 0.238 \\
    ENeRF [SIGGRAPH'22]~\cite{lin2022efficient}  & 28.90 & 0.967 & - & - & Ours & \textbf{24.28} & \textbf{0.959} & \textbf{0.101} \\
    \cmidrule(l{0pt}r{0pt}){6-9}
    \multicolumn{5}{c|}{} & \multicolumn{4}{c}{\textbf{HuMMan}} \\
    \multicolumn{5}{c|}{} & Method & PSNR↑ & SSIM↑ & LPIPS↓ \\
    \cmidrule(l{0pt}r{0pt}){6-9}
    HumanNeRF [arXiv'22]~\cite{weng2022humannerf}  & 30.66 & 0.969 & 0.033 & 10h / 0.30 & NHP[NIPS'21]~\cite{kwon2021neural} & 18.99 & 0.845 & 0.182 \\
    DVA [SIGGRAPH'22]~\cite{remelli2022drivable} & 29.45 & 0.956 & 0.038 & 1.5h / 16.5 & MPS-NERF [SIGGRAPH'22]~\cite{lin2022efficient} & 17.44 & 0.824 & 0.193 \\
    InstantNVR [CVPR'23]~\cite{geng2023learning}    & 31.01 & 0.971 & 0.038 & 5m / 2.20 & SHERF [ICCV'23]~\cite{Hu_2023_ICCV}& 20.83 & 0.891 & 0.125 \\
    InstantAvatar [CVPR'23]~\cite{jiang2023instantavatar}    & 29.73 & 0.938 & 0.068 & 3m / 4.15 & GHG [arXiv'24]~\cite{chen2024generalizable} & \textbf{23.86} & 0.952 & \textbf{0.0591} \\
    KeypointNeRF [ECCV'22]~\cite{mihajlovic2022keypointnerf}  & 25.03 & 0.898 & 0.104 & 20h / 1.05 & GST [arXiv'24]~\cite{prospero2024gst}& 18.40 & 0.87 & 0.14 \\
    SparseHumanNeRF [CVPR'22]~\cite{Zhao_2022_CVPR}    & 30.24 & 0.9679 & - & - & Ours & 22.18 & \textbf{0.977} & {0.060} \\
    \cmidrule(l{0pt}r{0pt}){6-9}
    \multicolumn{5}{c|}{} & \multicolumn{4}{c}{\textbf{DNA-Rendering}} \\
   \multicolumn{5}{c|}{} & Method & PSNR↑ & SSIM↑ & LPIPS↓ \\
    \cmidrule(l{0pt}r{0pt}){6-9}
    FlexNeRF [CVPR'23]~\cite{Jayasundara_2023_CVPR} & 31.73 & 0.9767 & 0.29 & - & HuGS [CVPR'24]~\cite{moreau2024human} & 31.5 & 0.98 & 0.022 \\
    MonoHuman [CVPR'23]~\cite{Yu_2023_CVPR}  & 30.05 & 0.9684 & 0.031 & - & DVA [SIGGRAPH'22]~\cite{remelli2022drivable} & 29.8 & 0.97 & 0.025 \\
    Sem-Human [arXiv'23]~\cite{zhang2023semantic}  & 30.80 & 0.967 & 0.033 & - & ENeRF [SIGGRAPH'22]~\cite{lin2022efficient} & 28.108 & 0.972 & 0.056 \\
    SMPLPix [WACV'21]~\cite{prokudin2021smplpix} & 27.00 & 0.91 & 0.090 & - &
    IBRNet [SIGGRAPH'23]~\cite{albahar2023single} & 27.844 & 0.967 & 0.081 \\
    GP-NeRF [ECCV'22]~\cite{chen2022gpnerf}  & 28.80 & 0.9408 & 0.090 & 20h / 1.05 & Im4D [SIGGRAPH'23]~\cite{lin2023high} & 28.991 & 0.973 & 0.062 \\
    AniNeRF [CVPR'21]~\cite{peng2021animatable} & 24.56 & 0.89 & 0.12 & - & 4K4D [CVPR'24]~\cite{Xu_2024_CVPR} & \textbf{31.173} & 0.976 & 0.055 \\
    INR [TPAMI'23]~\cite{peng2021neural}  & 30.54 & 0.970 & - & - & Ours & 30.18 & \textbf{0.989} & \textbf{0.012} \\
    \cmidrule(l{0pt}r{0pt}){6-9}
    \multicolumn{5}{c|}{} & \multicolumn{4}{c}{\textbf{People-Snapshot }} \\
    \multicolumn{5}{c|}{} & Method & PSNR↑ & SSIM↑ & LPIPS↓ \\
    \cmidrule(l{0pt}r{0pt}){6-9}
    HumanSplat [NIPS'24]~\cite{NEURIPS2024_87affd20}  & 29.82 & 0.9396 & 0.1048 & - &
     HumanNeRF [arXiv'21]~\cite{weng2022humannerf} & 26.90 & 0.9605 & 0.018 \\
     GPS-Gaussian [CVPR'24]~\cite{zheng2024gpsgaussian} & 29.68 & 0.95 & - & - & Neural Dressing [CVPR'21]~\cite{Grigorev_2021_CVPR} & - & 0.91 & 0.07 \\
    Deform3GS [SIGGRAPH'22]~\cite{lin2022efficient} & 24.10 & 0.869 & 0.126 & - & InstantAvatar [CVPR'23]~\cite{geng2023learning} & 29.53 & 0.9716 & 0.016 \\
    GoMAvatar [CVPR'24]~\cite{Wen_2024_CVPR} & 30.37 & 0.9689 & 0.032 & 23m / 4.15 & Anim-NeRF~ [arXiv'21]\cite{chen2021animatable} & 29.37 & 0.970 & 0.017 \\
     HuGS [CVPR'24]~\cite{moreau2024human}  & 26.58 & 0.934 & 0.022 & - & Neural Body [CVPR'21]~\cite{peng2021neural} & 25.49 & 0.928 & - \\
    SplatArmor [arXiv'23]~\cite{jena2023splatarmor} & 30.24 & - & 0.032 & - & GoMAvatar [CVPR'24]~\cite{Wen_2024_CVPR} & 30.68 & 0.9767 & 0.0213 \\
    3DGS-Avatar [CVPR'24]~\cite{qian20233dgsavatar}  & 30.61 & 0.9703 & - & 30m / 20 & SelfRecon [CVPR'22]~\cite{jiang2022selfrecon} & 24.91 & - & 0.061 \\
    Deform 3D [arXiv'23]~\cite{jung2023deformable}  & 29.28 & 0.964 & 0.040 & - & SMPLPix [WACV'21]~\cite{prokudin2021smplpix} & 17.90 & - & 0.165 \\
    HUGS [CVPR'24]~\cite{kocabas2024hugs}  & 30.54 & 0.970 & 0.030 & -  & FlexNeRF [CVPR'23]~\cite{Jayasundara_2023_CVPR}& 28.77 & 0.904 & 0.035 \\
    GauHuman [CVPR'24]~\cite{hu2024gauhuman}   & 31.34 & 0.965 & 0.031 & \textbf{1m / 189} & SplatArmor [arXiv'23]~\cite{jena2023splatarmor}& 27.08 & - & 0.43 \\
    GART [CVPR'24]~\cite{Lei_2024_CVPR} & \textbf{32.22} & 0.977 & 0.29 & - & GART [CVPR'24]~\cite{Lei_2024_CVPR}& 30.40 & 0.976 & 0.037 \\
    Ours & 30.86 & \textbf{0.979} & \textbf{0.028} & 30m / 100 & Ours &  \textbf{32.86} & \textbf{0.98} & \textbf{0.014} \\
    \bottomrule
  \end{tabular}
  }
  \caption{Comparison of methods on ZJU-MoCap \cite{peng2021neural}, Twindom\cite{twindom}, HuMMan\cite{cai2022humman}, DNA-Rendering\cite{cheng2023dna} and People-Snapshot \cite{alldieck2018video} datasets. }
  \label{tab:merged_comparison}
\end{table*}

{\bf System Setup for Runtime Performance } 
All experiments were conducted on a workstation with an AMD Ryzen 9 5900X CPU, 64 GB RAM, and an NVIDIA RTX 4080 GPU (16 GB VRAM). All input images are resized to $512 \times 512$ pixels, and our pipeline achieves interactive performance with a per-frame processing time of approximately 100 milliseconds. The end-to-end processing is decomposed as follows: the Gaussian Splatting (GS) rendering module consumes $\sim$38 ms; pose feature extraction using a lightweight pose encoder takes $\sim$12 ms; image feature extraction via a ResNet-based backbone requires $\sim$20 ms; and our Temporal Pose Stabilizer (TPS), which adjusts view blending based on inter-frame pose dynamics, contributes $\sim$18 ms. The remaining $\sim$12 ms accounts for view-dependent shading, memory overhead, and post-processing (e.g., alpha blending and tone mapping).

{\bf Model Specification: } We build our method upon a U-Net architecture tailored to multi-scale feature preservation and uncertainty-aware prediction. All training configurations, implementation details, and hardware specifications are provided in the \textit{Appendix}. Our model uses three parallel encoders, each beginning with a $3 \times 3$ convolution (32 channels), followed by six residual units with progressively increasing widths (32, 64, 96, 128). Each residual block incorporates a Squeeze-and-Excitation (SE) module for channel-wise attention~\cite{Hu_2018_CVPR}. Skip connections align encoder and decoder features to preserve multi-scale information. The decoder upsamples the features and outputs three heads: (1) an SR-Opacity head predicting Gaussian parameters—scale (3 channels, Softplus), rotation (3 channels, normalized), and opacity (2 channels, Sigmoid); (2) a depth head for auxiliary supervision; and (3) a confidence head for per-pixel uncertainty. The final prediction $\hat{\mathbf{G}}(x)$ is a blend of the predicted output $\mathbf{G}_{\text{dec}}(x)$ and prior $\mathbf{G}_{\text{prior}}(x)$, modulated by the confidence maps $\mathbf{c}(x)$:
\begin{equation}
\hat{\mathbf{G}}(x) = \sigma(\mathbf{c}(x)) \cdot \mathbf{G}_{\text{dec}}(x) + (1 - \sigma(\mathbf{c}(x))) \cdot \mathbf{G}_{\text{prior}}(x)
\end{equation}
where $\mathbf{c}(x)$ is a learned confidence map. A broader discussion of uncertainty-aware blending strategies can be found in~\cite{kendall2017uncertainties}.

\subsection{Comparison to State-of-the-arts}
We evaluate \textit{PoseGaussian} on several public benchmarks for dynamic human reconstruction, including controlled multi-view studio datasets (e.g., ZJU-MoCap) and large-scale benchmarks with diverse poses and motions (e.g., THuman2.0, HuMMan, DNA-Rendering).

\begin{figure}[t]
  \centering
  \includegraphics[width=\linewidth]{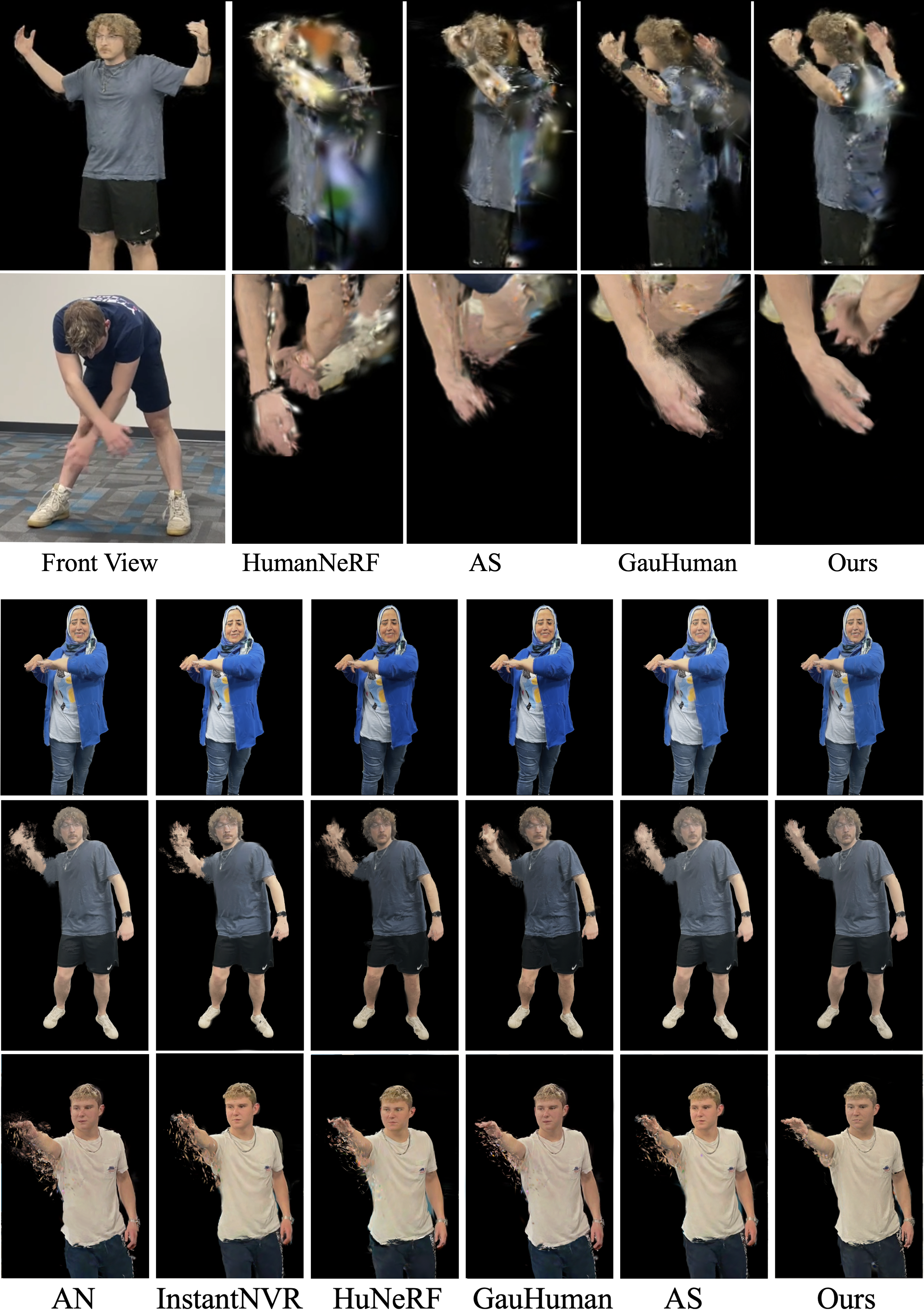}
  \caption{Challenging cases in pose synthesis. Top: occlusion scenario revealing difficult views of hidden regions (e.g., back, inner arms). Bottom: fast motion scenario comparing recent NeRF- and Gaussian-based methods (see Fig.~\ref{fig:teaser} for references)
}
  \label{fig:challenges}
\end{figure}

{\bf Baseline Evaluation Protocols and Setup Standardization}  
To ensure fair comparison across baselines with differing protocols, we re-implemented or adapted methods to match our pipeline, using official pretrained models or public code when available. The testing setup was unified by fixing 8 input views uniformly sampled across the rig, unlike HumanNeRF~\cite{weng2022humannerf} (which omits Camera 1) and InstantNVR~\cite{geng2023learning} (which omits Camera 4). All methods used the same temporal sequences and test splits, and evaluation metrics were computed from the same view across methods for alignment; when pretrained models could not be re-evaluated due to dependency or training complexity, we reported published results verbatim and noted protocol differences. For baselines lacking perceptual metrics such as LPIPS, we attempted to re-run their released models; when not feasible, we marked metrics as unavailable. Since baselines differ in input modality—HumanNeRF~\cite{weng2022humannerf}, InstantNVR~\cite{geng2023learning}, and MonoHuman~\cite{Yu_2023_CVPR} use monocular video, while GPS-Gaussian~\cite{zheng2024gpsgaussian} and NHP~\cite{kwon2021neural} rely on multi-view images—we adapted monocular methods by extracting single-camera trajectories from our multi-view dataset, ensuring compatibility.

\textbf{Quantitative Benchmark Comparison:} Table~\ref{tab:merged_comparison} presents a comprehensive quantitative evaluation of our method against state-of-the-art approaches across a diverse set of public benchmark datasets. By leveraging pose as guidance, our method consistently achieves superior structural consistency, attaining the highest SSIM scores across all datasets---including a peak score of 0.979 on ZJU-MoCap~\cite{peng2021neural}. The integration of structural cues and temporal consistency further enhances perceptual quality, as reflected by favorible LPIPS scores ranging from 0.014 to 0.101, underscoring the method's effectiveness in preserving fine appearance details and visual coherence. While our method did not achieve the highest PSNR, it remains competitive, which is reasonable given that PSNR primarily measures pixel-wise differences and may not fully capture perceptual or temporal quality improvements. In addition, Table~\ref{tab:RealWorldDataset} focuses exclusively on the THuman 2.0 dataset, where \textit{6-view} and \textit{8-view} settings on selected sequences are employed to construct controlled benchmarks. This targeted analysis enables a more rigorous evaluation under varying view configurations, which is essential for assessing pose-guided novel view synthesis. Across all tested conditions on THuman 2.0, our method consistently outperforms competing methods, achieving an SSIM score of 0.957 and an LPIPS score of 0.031, demonstrating strong generalization and rendering quality.

\begin{table*}[t]
  \centering
  \resizebox{\textwidth}{!}{%
  \begin{tabular}{lccc|ccc|ccc|ccc}
    \toprule
    \multicolumn{13}{c}{\textbf{THuman2.0 Dataset}} \\
    \cmidrule(r){2-13}
    \multicolumn{1}{c}{} & \multicolumn{6}{c|}{\textbf{Training}} & \multicolumn{6}{c}{\textbf{Testing}}  \\
    \cmidrule(r){2-7} \cmidrule(lr){8-13} \cmidrule(l){11-13}
    Method & \multicolumn{3}{c|}{6-camera setup} & \multicolumn{3}{c|}{8-camera setup} & \multicolumn{3}{c|}{Real World Data} & \multicolumn{3}{c}{Rendered Data} \\
    & PSNR↑ & SSIM↑ & LPIPS↓ & PSNR↑ & SSIM↑ & LPIPS↓ & PSNR↑ & SSIM↑ & LPIPS↓ & PSNR↑ & SSIM↑ & LPIPS↓ \\
    \midrule
    3D-GS [ACMTOG’23]~\cite{kerbl2023gaussian} & - & - & - & - & - & - & 22.97 & 0.839 & 0.125 & 24.18 & 0.821 & 0.144 \\
    FloRen [SIGGRAPH'22]~\cite{shao2022floren} & 18.72 & 0.770 & 0.267 & 23.26 & 0.812 & 0.184 & 22.80 & 0.872 & 0.136 & 23.26 & 0.812 & 0.184 \\
    IBRNet [CVPR'21]~\cite{Wang_2021_CVPR} & 21.08 & 0.790 & 0.263 & 23.38 & 0.836 & 0.212 & 22.63 & 0.852 & 0.177 & 23.38 & 0.836 & 0.212 \\
    NARF [SIGGRAPH'22]~\cite{lin2022efficient} & - & - & - & - & - & - & 21.80 & 0.8088 & - & - & - & - \\
    PIFu [ICCV'19]~\cite{saito2019pifu} & - & - & - & - & - & - & - & - & - & 20.40 & 0.921 & 0.079 \\
    ENeRF [SIGGRAPH'22]~\cite{lin2022efficient} & 21.78 & 0.831 & 0.181 & 24.10 & 0.869 & 0.126 & 23.26 & 0.893 & 0.118 & 24.10 & 0.869 & 0.126 \\
    SHERF  [ICCV’23]~\cite{Hu_2023_ICCV} & - & - & - & - & - & - & 24.26 & 0.91 & 0.11 & - & - & - \\
    Contex-Human [CVPR'24]~\cite{gao2024contex} & - & - & - & - & - & - & - & - & - & 21.40 & 0.923 & 0.063 \\
    DoubleField [CVPR'22]~\cite{Shao_2022_CVPR} & - & - & - & - & - & - & 25.10 & 0.905 & -- & - & - & - \\
    GPS-Gaussian [CVPR’24]~\cite{zheng2024gpsgaussian} & 23.03 & 0.884 & 0.168 & 25.57 & 0.898 & 0.112 & 24.64 & 0.917 & 0.088 & 25.57 & 0.898 & 0.112 \\
    FreeSplat [NIPS'24]~\cite{NEURIPS2024_c2166d01} & 23.35 & 0.843 & 0.184 & -- & -- & -- & 25.90 & 0.808 & 0.252 & - & - & - \\
    RoGSplat[arXiv'25]~\cite{xiao2025rogsplat} & 23.12 & 0.8980 & 0.1661 & -- & -- & -- & \textbf{25.99} & 0.9452 & 0.057 & - & - & - \\
    HumanSplat [NIPS’24]~\cite{NEURIPS2024_87affd20} & - & - & - & - & - & - & - & - & - & 24.033 & 0.918 & 0.055 \\
    LIFe-GoM [arXiv'25]~\cite{wen2025life} & - & - & - & 24.65 & - & 0.110 & 25.32 & -- & 0.099 & - & - & - \\
    SIFU [CVPR'24]~\cite{zhang2024sifu} & - & - & - & - & - & - & - & - & - & 22.102 & 0.923 & 0.079 \\
    HumanSGD [SIGGRAPH'23]~\cite{albahar2023single} & - & - & - & - & - & - & - & - & - & 17.365 & 0.895 & 0.130 \\
    TeCH [3DV'24]~\cite{huang2024tech} & - & - & - & - & - & - & - & - & - & 25.211 & 0.936 & 0.083 \\
    \textbf{Ours} & \textbf{24.9} & \textbf{0.94} & \textbf{0.09} & \textbf{26.07} & \textbf{0.927} & \textbf{0.081} & 25.47 & \textbf{0.966} & \textbf{0.05} & \textbf{25.78} & \textbf{0.957} & \textbf{0.031}  \\
    \bottomrule
  \end{tabular}
  }
  \caption{Comparison of methods on the THuman2.0 test set using 6-camera and 8-camera setups, covering both Real World and Rendered data.}
  \label{tab:RealWorldDataset}
\end{table*}

{\bf Qualitative Comparison:}
We qualitatively compare our approach under two challenging scenarios: self-occlusion and fast motion. The top rows of Fig.~\ref{fig:challenges} illustrate the self-occlusion scenario, where virtual views are rendered from viewpoints significantly different from the front-facing input cameras (shown in the first column of each row), exposing heavily occluded regions such as the back of the person in Scene~1 and the inner arms in Scene~2. The zoomed-in novel views demonstrate that our method better preserves sharp and continuous body contours---particularly along the challenging back and shoulder regions---and maintains structural details such as clear arm separation, whereas competing methods exhibit noticeable blurring and fragmentation. The bottom rows of Fig.~\ref{fig:challenges} evaluate reconstruction quality under varying motion speeds using only six input views to highlight method differences: \textit{Row 1} (slow motion) shows subtle differences, while faster motions in \textit{Rows 2} and \textit{3} reveal artifacts such as floaters around fast-moving hands and arms. Overall, our method produces more stable and coherent results, although minor distortions may still occur under very rapid motion. Many artifacts are significantly reduced with more input views; we deliberately chose a sparse setup to emphasize the challenge and demonstrate robustness, while denser captures yield near artifact-free results.

{\bf Temporal Coherence and Motion Stability:} 
To demonstrate the role of pose in enhancing temporal coherence, we conduct a comprehensive evaluation using SSIM, PSNR, and LPIPS. Here, we present SSIM as a representative example. Our evaluation covers 32 motion sequences featuring challenging dynamics such as rapid limb movement. For statistical robustness, we analyze the average and standard deviation of per-frame $\Delta$SSIM, as well as $\Delta$SSIM values averaged across all sequences. As shown in Table~\ref{tab:temporal_consistency}, our method achieves the lowest $\mu(\Delta\text{SSIM})$ of 0.023 and $\sigma(\Delta\text{SSIM})$ of 0.015, almost half of the next best method. This nearly 50\% reduction in frame-to-frame SSIM variation translates to visibly smoother and more temporally stable renderings, effectively suppressing the jitter and flickering artifacts commonly observed in competing methods.

\begin{table}[t]
\centering
\resizebox{0.48\textwidth}{!}{%
\begin{tabular}{l|c|c}
\toprule
\textbf{Method} & \(\mu(\Delta\text{SSIM})\) & \(\sigma(\Delta\text{SSIM})\) \\
\midrule
3DGS \cite{kerbl2023gaussian} & 0.052 & 0.039 \\
IBRNet \cite{Wang_2021_CVPR} & 0.0451 & 0.035 \\
GPS-Gaussian \cite{zheng2024gpsgaussian} & 0.040 & 0.028 \\
\textbf{Ours} & \textbf{0.023} & \textbf{0.015} \\
\bottomrule
\end{tabular}%
}
\caption{Frame-to-frame $\Delta$SSIM consistency.}
\label{tab:temporal_consistency}
\vspace{-1em}
\end{table}

\subsection{Ablation Study}

\begin{figure}[t]
  \centering
  \includegraphics[width=0.48\textwidth]{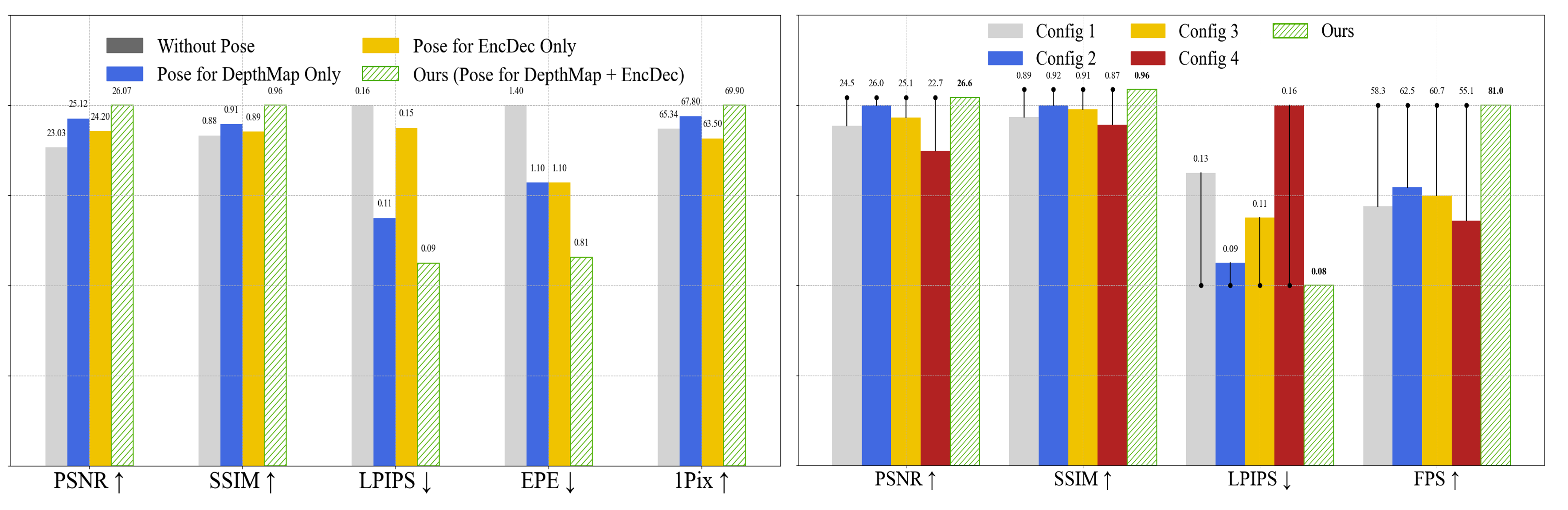}  
  \caption{(Left) Impact of pose information. (Right) Impact of pose encoder configurations.}
  \label{fig:ablation_params}
\end{figure}

{\bf Fusion Strategies: } 
We explored several fusion strategies to combine the outputs \( \{\mathbf{f}_{img}^{(s)}\}^{s \in \mathcal{S}} \) and \( \{\mathbf{f}_{pose}^{(s)}\}^{s \in \mathcal{S}} \) from Eq. \ref{eq:poseandfeature}, including concatenation, Feature-wise Attention \cite{zhang2021multimodal}, and Gated Fusion \cite{arevalo2017gated}. Feature-wise Attention achieved the best performance, with the lowest EPE (0.75) and 1px accuracy of 78\%, at a moderate model size (1.2M parameters). Concatenation offered a favorable trade-off between accuracy and complexity (EPE 0.80, 1px 80\%, 0.8M parameters), while Element-wise Addition provided a simpler, more lightweight option (0.5M parameters) but with higher EPE (1.00). Weighted Average Fusion \cite{song2009novel} and Element-wise Multiplication also delivered reasonable compromises (EPE \(\sim\)1.1 and 0.98; 0.9–1.0M parameters). More complex methods such as Outer Product Fusion \cite{terra2019spectral} increased parameter counts substantially (1.3–1.8M) with only marginal improvements. Based on these results, we select concatenation as the best balance between accuracy and efficiency.

\begin{table}[h]
\centering
\begin{tabular}{cccccc}
\toprule
\(\beta\) & \(\gamma\) & \(\lambda\) & PSNR & SSIM & LPIPS \\
\midrule
0.5 & 0.5 & 0.0 & 19.9 & 0.879 & 0.205 \\
0.3 & 0.7 & 0.0 & 19.6 & 0.871 & 0.212 \\
0.8 & 0.2 & 0.0 & 20.3 & 0.865 & 0.225 \\
0.5 & 0.5 & 0.1 & 21.7 & 0.890 & 0.180 \\
0.5 & 0.5 & 0.5 & 25.1 & 0.905 & 0.148 \\
0.5 & 0.5 & 1.0 & 24.9 & 0.940 & 0.090 \\
\bottomrule
\end{tabular}
\caption{Ablation study of loss hyperparameters \(\beta\), \(\gamma\), and pose fusion weight \(\lambda\) on the THuman2.0 dataset.}
\label{tab:ablation_loss}
\end{table}

\noindent
\textbf{Loss Hyperparameters:} 
We analyze the effect of the loss weights \( \beta \), \( \gamma \), and the pose fusion factor \( \lambda \) through an ablation study, as shown in Table~\ref{tab:ablation_loss}. In the absence of pose fusion (\( \lambda = 0 \)), a balanced configuration of RGB and depth losses (\( \beta = \gamma = 0.5 \)) achieves better performance than skewed settings, which trade off perceptual quality for structural accuracy or vice versa. Introducing pose fusion (\( \lambda > 0 \)) consistently improves performance across all metrics. As \( \lambda \) increases, we observe steady gains, with the best results obtained when full pose fusion is applied (\( \lambda = 1.0 \)) along with balanced loss weights.

{\bf Pose and Architecture Effects:} 
Fig.~\ref{fig:ablation_params} illustrates the effects of pose feature integration (left) and network architecture variations (right) on performance. Integrating pose inputs into both the depth map and encoder--decoder pathways delivers the best results, significantly outperforming configurations that incorporate pose in only one component or omit it altogether. This integration notably enhances perceptual quality, improving the LPIPS score by 45\% (0.09 vs. 0.16) compared to the model without pose information. For network design, we evaluated four alternative architecture variants---detailed in the Appendix---that explore the use of Batch Normalization, Residual Blocks, and downsampling layers. Their performance is summarized in Fig.~\ref{fig:ablation_params} (right). These ablation studies unequivocally demonstrate that the integration of pose information at both the geometric and feature-encoding stages is critical to the superior performance of PoseGaussian.

\section{Conclusion} We demonstrate that leveraging pose as a structural prior combined with temporal constraints enables our method to achieve state-of-the-art performance on standard benchmarks across multiple public datasets. PoseGaussian effectively reduces motion-induced artifacts in dynamic human scenes, resulting in superior temporal coherence and visual stability in novel view synthesis. While our approach significantly improves reconstruction quality under challenging fast motions, some artifacts persist, especially when the virtual viewpoint shifts considerably from the original capture. Future work will extend our framework to handle multiple interacting people, addressing challenges in occlusion, interaction modeling, and temporal consistency to further enhance reconstruction fidelity in complex dynamic scenes.

{\small
\bibliographystyle{ieee_fullname}
\bibliography{egbib}
}

\end{document}